\pdfoutput=1
\documentclass[review]{elsarticle}
\usepackage{subfigure}
\usepackage{longtable}
\usepackage{algorithm, algorithmic}
\usepackage{bm}
\usepackage{booktabs}
\usepackage{multirow}
\usepackage{geometry}
\usepackage{amsfonts,amssymb}
\usepackage{float}

\usepackage{lineno,hyperref}
\modulolinenumbers[5]

\journal{arXiv}

\begin{document}
\begin{frontmatter}
\title{Bilevel Online Deep Learning in Non-stationary Environment}

\author[mymainaddress]{Ya-nan Han}

\author[mymainaddress]{Jian-wei Liu\corref{mycorrespondingauthor}}
\cortext[mycorrespondingauthor]{Corresponding author}
\ead{liujw@cup.edu.cn}

\author[mymainaddress]{Bing-biao Xiao}

\author[mymainaddress]{and Xiong-lin Luo}

\address[mymainaddress]{Department of Automation, College of Information Science and Engineering, China University of Petroleum, Beijing Campus (CUP), Beijing, China}
\begin{abstract}
Recent years have witnessed enormous progress of online learning. However, a major challenge on the road to artificial agents is concept drift, that is, the data probability distribution would change where the data instance arrives sequentially in a stream fashion, which would lead to catastrophic forgetting and degrade the performance of the model. In this paper, we proposed a new Bilevel Online Deep Learning (BODL) framework, which combine bilevel optimization strategy and online ensemble classifier. In BODL algorithm, we use an ensemble classifier, which use the output of different hidden layers in deep neural network to build multiple base classifiers, the important weights of the base classifiers are updated according to exponential gradient descent method in an online manner. Besides, we apply the similar constraint to overcome the convergence problem of online ensemble framework. Then an effective concept drift detection mechanism utilizing the error rate of classifier is designed to monitor the change of the data probability distribution. When the concept drift is detected, our BODL algorithm can adaptively update the model parameters via bilevel optimization and then circumvent the large drift and encourage positive transfer. Finally, the extensive experiments and ablation studies are conducted on various datasets and the competitive numerical results illustrate that our BODL algorithm is a promising approach.
\end{abstract}
\end{frontmatter}

\section{Introduction}

Deep learning techniques have achieved enormous success in a wide range of artifi-cial intelligence (AI) and machine learning applications in recent years\cite{1bengio2013representation,2chen2015net2net}. However, most of these existing deep learning approaches suppose that the models often work in a batch learning setting or offline learning fashion, where the entire training dataset must be available to train a model by some learning techniques. Such learning ap-proaches are poorly scalable for many real-word tasks, where the data instances arrive in a sequential manner. Thus, making deep learning available for the streaming data is a desideratum in the field of machine learning.\\
Unlike traditional batch learning, online learning represents a significant family of learning algorithms that are designed to optimize and learn models incrementally over streaming data sequentially\cite{3cesa2006prediction}. Online learning shows the tremendous ad-vantages that the models can be updated efficiently in an online manner compared with traditional offline learning fashion when the new data instance comes. Similar to batch learning algorithms, online learning can also be applied for various real-word tasks, such as supervised classification task\cite{4sahoo2018online}, unsupervised learning task\cite{5hoi2018online}, and so on. \\
However, in general, online learning algorithms cannot be directly employed to deep neural network. They have to cope with the intractable convergence problems, such as vanishing gradient. Besides, the traditional shallow or fixed neural network structure is poorly scalable for the most real-world applications where the data in-stances arrive in a sequential order and the probability distribution of data is non-stationary. Therefore, a promising online deep learning framework should be devel-oped that can effectively and rapidly learn knowledge in non-stationary.\\
It should also be noted that the probability distribution obeyed by streaming data could occur the concept drift, in other words, the data probability distribution chang-es. In this circumstance, the leaning algorithms must take some actions to prevent the large drift and encourage positive transfer, in other words, the learner should make a trade-off between both the new and old knowledge and alleviate the catastrophic forgetting. The classical algorithms for catastrophic forgetting are Elastic Weight Consolidation (EWC)\cite{6kirkpatrick2017overcoming} and their variants\cite{7chaudhry2018riemannian}, but this kind of algorithms attempt to address catastrophic forgetting by augmenting objective function and then control the whole network, that is, let the learning model’s weights balance between these two factors, rather than directly take actions to cope with catastrophic forgetting. Based on the above fact, therefore this reminds us of the importance to enhance the differ-ent-depth latent representations and the ability to rapidly adapt to dynamic changing situations.\\
To achieve this, in this work, we devise a novel Bilevel Online Deep Learning (BODL) framework, which consists of three major components: online ensemble classifier, concept drift detection and bilevel online deep learning. Our BODL frame-work can effectively utilize the different abstract level latent feature representations to build classifiers via the online ensemble framework, where the important weights of the base classifiers would be updated by online exponential gradient descent strat-egy. consider the convergence problem of online ensemble framework, we apply the similar constraint to generate the favorable latent representation. Besides, a concept drift detected mechanism is devised according to the error rate of base classifiers. When the concept drift is detected, our BODL model can adaptively update the model parameters via bilevel optimization and then prevent the large drift and en-courage positive transfer. \\
In a summary, our main contributions in this paper are listed below: \\
1) We design an effective bilevel learning strategy. Specifically, if the concept drift is detected, the model would adaptively adjust the parameters   for all base classi-fiers and   of the different-depth feature representation mentioned in section 2 using bilevel optimization, where this process is achieved based on a tiny episodic memory. After that, the model can circumvent the large drift and encourage positive transfer in non-stationary environment.\\
2) In this work, consider the convergence problem of online ensemble framework, we impose the similar constraint between the shallower and the deeper layer’s fea-ture, which would be beneficial to generate the favorable feature representations.\\
3) The comparative experiments are devised to verify the effectiveness of the pro-posed BODL algorithm, and we analysis the experimental results of a variety of algo-rithms from different perspectives in terms of accuracy, precision, recall-score and F-1 score, and then we can see that our BODL algorithm can exploit the different-depth feature representations and adapt to rapidly changing environment.\\
The remainder of this paper is organized as follows. In Section 2, we introduce our BODL algorithm in details, which consists of three parts: online ensemble classifier, concept drift detection mechanism, bilevel learning for concept drift. In Section 3 we empirically compare BODL algorithm with several state-of-the-art online learning algorithms. In Section 4 we elaborate related works. In Section 5 we summarize the whole work and the interested directions in the future.

\section{Bilevel Online Deep Learning(BODL)}

In this work, we present bilevel online deep learning, a conceptually novel framework for online learning based on bilevel optimization\cite{8jenni2018deep} and online ensemble framework. Our BODL architecture can be divided into three main parts: online ensemble classi-fier, concept drift detection mechanism, bilevel learning for concept drift. The online deep ensemble classifier can make a trade-off among the different-level base classi-fiers and improve the performance of classification; Concept drift detection mecha-nism is used to monitor the change in non-stationary environment; When the concept drift is detected, bilevel learning is designed to adaptively adjust the parameters $\theta _n^t$ and $W_n^t$ , then the model can adapt to the change in non-stationary environment.

\subsection{Online Ensemble Classifier}

We illustrate the online deep ensemble classifier in Fig.1, where ${\omega ^t} = [\omega _1^t,...,\omega _N^t]$ represents the importance of the N base classifiers. The online deep ensemble classifi-er can make a trade-off among the different-level base classifiers via Exponential Gradient Descent(EGD) algorithm in an online manner\cite{4sahoo2018online}.\\
More specifically, we character a Deep Neural Network (DNN) with   hidden layers, and the final ensemble classifier can be achieved by dynamically updating the weight parameters of the base classifiers for each hidden layer based on their classifi-cation loss. The specific ensemble prediction function can be written as Eq.(1).
\begin{equation}
	\begin{array}{c}
	\label{(1)}
	F(x) = \sum\limits_{n = 0}^N {{\omega _n}{f_n}}\\
	{f_n} = soft\max ({h_n}{\theta _n}),\forall n = 0,...,N\\
	{h_n} = \sigma ({W_n}{h_{n - 1}}),\forall n = 1,...,N\\
	\end{array}
\end{equation}

Compared to the traditional network, in which the feature representation con-structed by outputs of the final hidden layer is used as input of the classifier, here we can make a favorable classifier by an online ensemble framework, which can benefit from the different depth feature representation and improve the prediction perfor-mance of the whole model. It is noted that the parameters $\omega _n^t$ ,$\theta _n^t$ and $W_n^t$ in Eq.(1) can be learned in an online flavor.
\begin{figure}[!htbp]
	\centering
	\includegraphics[scale=1.0]{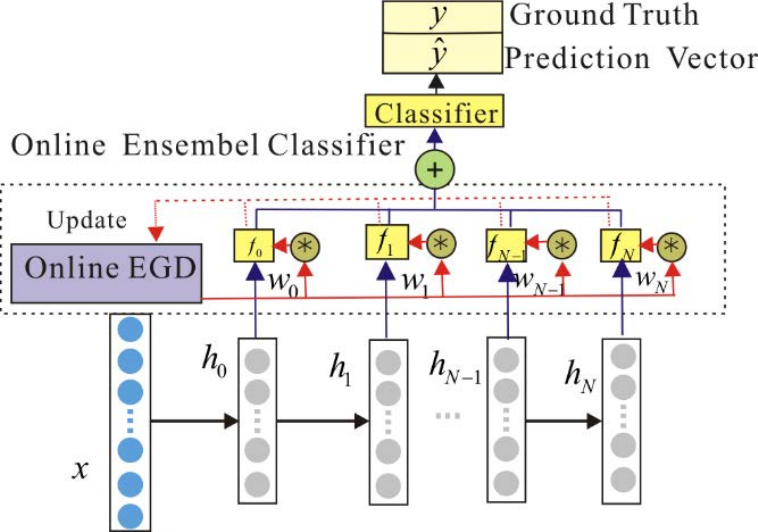}
	\caption{Online deep ensemble classifier.}
	\label{fig1}
\end{figure}

\textbf{Update the parameters} ${\omega _n}$ We update the weights ${\omega _n}$ for base classifiers using exponential gradient descent11. Firstly, the weights ${\omega}$ are initialized using a uniform distribution:${\omega _n} = \frac{1}{{N + 1}},n = 0,...,N$ i.e., each base classifier has equal probability to be picked. At each iteration, the prediction loss of the n-th base classifier ${f_n}$  can be written as ${\cal L}(\hat y_n^t,y_n^t)$ , where   $\hat y_n^t$ and $y_n^t$  represent the base classifier prediction and the target variable respectively. Then, the weight of each base classifier can be learned according to the loss suffered and the update rule is given by follow:
\begin{equation}
	\label{(2)}
	\omega _n^{t + 1} \leftarrow \omega _n^t{e^{ - \eta {\cal L}(\hat y_n^t,y_n^t)}}
\end{equation}
where $\eta  \in (0, + \infty )$  and $\eta$  is set to 0.01 in our work. After that, the trained base clas-sifier’s important weight is discounted by an exponential weight $e^{ - \eta {\cal L}(\hat y_n^t,y_n^t)}$. \\
\textbf{Update the parameters}  ${\theta _n^t}$   The parameters ${\theta _n^t}$  for all base classifiers are updated using Stochastic Online Gradient Descent (SOGD), and this process is analogical to the traditional feedforward networks.\\
\textbf{Update the parameters}  $W_n^t$   The update rule about the parameters $W_n^t$  of the dif-ferent-depth feature representation is different from the traditional backpropagation framework. The objective function includes two parts: the adaptive loss function and similar constraint, which are defined as follow:
\begin{equation}
	\label{(3)}
	{\cal L}(F(x),y) = \sum\limits_{n = 0}^N {{\omega _n}} {{\cal L}_{pre}}({f_n}(x),y) + \lambda {{\cal L}_{sim}}({h_{se}},{h_{de}})
\end{equation}
where, the first part in loss function represents the adaptive prediction loss. Note that, the parameters of shallower layer tend to converge faster than the ones of deeper layer, which can lead to deeper base classifiers learn slowly\cite{2chen2015net2net}. Thus, we incorporate the similar constraint between the shallower and deeper layer’s features, which can be beneficial to generate the favorable feature representations and improve the conver-gence rate and the prediction performance of the deeper layer. In this work, $\lambda$  is a tradeoff parameter and is set to 0.1. Note that, the similarity can be modelled in mul-tiple manners and we choose the squared distance metric in this paper.

\subsection{Bilevel Online Deep Learning}
As the streaming data comes gradually and the data probability distribution could change. We monitor the change of the data probability distribution utilizing the error rate of classifier. This concept drift detection mechanism is similar to the drift detec-tion method in \cite{10kifer2004detecting} but the warning phase is not arranged in this paper in order to avoid the use of slide window methods. In this section, we describe our adaptive online deep learning based on bilevel optimization in detail. Figure 2 shows a flowchart of the bilevel online deep learning framework.
\begin{figure}[!htbp]
	\centering
	\includegraphics[scale=1.0]{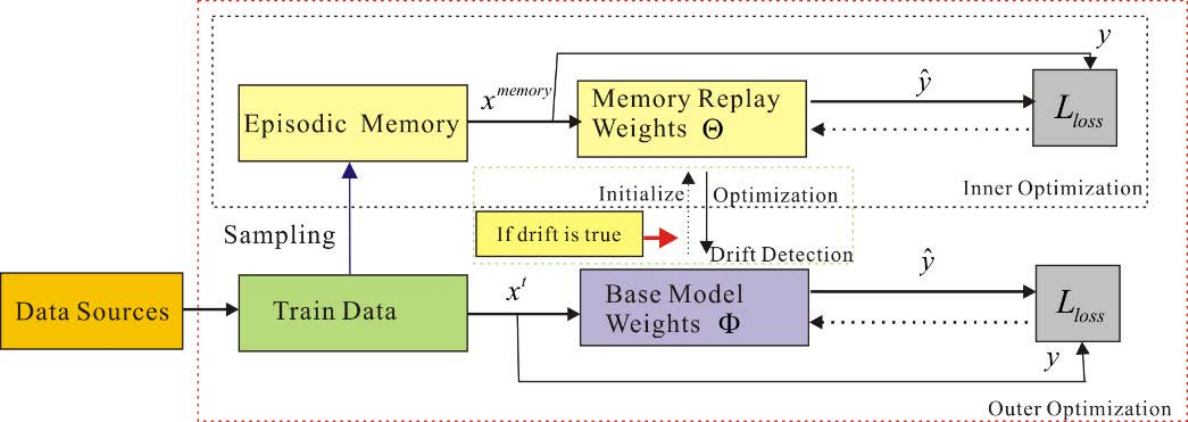}
	\caption{The bilevel online deep learning framework. BODL utilizes the memory replay weights $\Theta $  and the episodic memory to optimize the base classifiers’ weights $\Phi $   in the non-stationary environ-ment, where the episodic memory is obtained by reservoir sampling.}
	\label{fig2}
\end{figure}

\subsubsection{Bilevel Learning}
For each arriving instance in online learning scenario, we detect the concept drift uti-lizing the error rate of classifier. If the concept drift is observed, the learning algorithm obviously needs to takes some actions to prevent large drift and achieve online in-cremental learning. Specifically, when the concept drift occurs, BODL initializes a memory weight $\Theta $  to replay the knowledge in the memory. Then we apply the trained memory weight $\Theta^* $  to update $\Phi $  such that it can prevent large drift and weight the new and old knowledge in a non-stationary environment as shown in the Fig.2. \\
Bearing this in mind, the objective function can be defined as the following bilevel optimization problem:
\begin{equation}
	\begin{array}{c}
	\mathop {\min }\limits_\Phi  {{\cal L}^{outer}}(\Theta *(\Phi ),\beta _{train}^t;\Phi )\\
	s.t.{\Theta ^*} = \arg \mathop {\min }\limits_\Theta  {{\cal L}^{inner}}({\Theta ^*},{{\cal M}^{memory}})
	\label{(4)}
	\end{array}
\end{equation}
where $\beta _{train}^t$  denotes the current training data that exists concept drift. We parame-terize each $\Phi$  as an inner optimization problem ${\cal L}_{\Phi}^{memory}$ , which the learner optimiz-es the corresponding $\Theta$ . During the bilevel learning, firstly the agent learns the memory weight $\Theta^*$  about the inner problem. After that, the agent learns the outer problem with respect to $\Phi$ . In this process, we apply the cross-entropy loss as objec-tive function for the inner and outer problems respectively.

\subsubsection{First Order Approximation}
Generally, the data comes gradually in non-stationary environment and the concept drift mechanism will monitor the change in online manner. When the concept drift occurs, the learner can adaptively adjust the model parameters and weight the new and old knowledge in a non-stationary environment via bilevel learning. \\
Specifically, assume that for an incoming training data $\beta _{train}^t$  reported as concept drift, the inner problem is settled by:
\begin{equation}
	{\Theta _i} \leftarrow {\Theta _i} - \mu {\nabla _{{\Theta _i}}}{\cal L}({\Theta _i},\beta _{train}^t;\Phi )where{\Theta _0} \leftarrow \Phi 
	\label{(5)}
\end{equation}
After receiving $\Theta^*$  via Eq. (5), the outer learning for the parameters $\Phi*$  can be solved by the chain rule.
\begin{equation}
	\begin{array}{c}
		{\Phi _t} \leftarrow {\Phi _t} - \gamma {\nabla _\Phi }{\cal L}({\Theta ^*},{{\cal M}^{memry}})\\
 		\leftarrow {\Phi _t} - \gamma \frac{{\partial {\Theta ^*}}}{{\partial \Phi }} \cdot \frac{\partial }{{\partial {\Theta ^*}}}{\cal L}({\Theta ^*},{{\cal M}^{memry}})
	\end{array}
	\label{(6)}
\end{equation}
Note that solving the Eq. (6) is a cumbersome problem in real word scenario be-cause of the Hessian vector product in the second term\cite{11pham2020bilevel}. In order to improve the efficiency of the computation, we apply first-order approximation to simplify the Eq.(6) in this work\cite{12nichol2018first,13zhang2019lookahead}. Thus, the outer learning is given by interpolating only in the parameter space:
\begin{equation}
	\begin{array}{c}
	{\Phi _t} ={\Phi _t} + \gamma ({\Theta ^{\prime}} - {\Phi _t})\\
	where{\Theta ^{\prime}} = {\Theta ^*} - \mu {\nabla _{{\Theta ^*}}}{\cal L}({\Theta ^*},{{\cal M}^{memory}})
	\end{array}
	\label{(7)}
\end{equation}
We apply Eq.(7) to obtain a one-step look-ahead parameter $\Theta ^{\prime}$  from $\Theta ^*$ . After that, we can adjust $\Phi$  by linearly interpolate between the current parameters  $\Phi$  and $\Theta ^{\prime}$  . It is noted that we only maintain the parameters of the main model $\Phi$,i.e., once the parameters $\Phi$  is obtained and then we discard it after every outer update. In this process, the inner optimization should be carried out via tiny experience memory\cite{14chaudhry2019continual}. 
\subsubsection{Bilevel Online Deep Learning Algorithm}
In this section, we show that our BODL algorithm can effectively learn in non-stationary environment by an online manner.\\
Our proposed BODL algorithm is shown in \textbf{Algorithm 1}.
\begin{algorithm}
	\caption{Bilevel Online Deep Learning algorithm(BODL)}
	\label{alg1}
	\begin{algorithmic}
		\STATE	\textbf{Input:}Discounting coefficient $\eta  \in (0, + \infty )$ ; Learning Rates: $\mu$,$\gamma$  ,  
		\STATE	\textbf{Require:}Memory management strategy for  ${\cal M}^{memry}$
		\STATE	\textbf{Initialize:} $F(x) = DNN$ with $N + 1$ hidden layers;$\omega _n^t = \frac{1}{{N + 1}},\forall n = 0,...,N;\Phi ;$
		\STATE ${\cal M}^{memry}\leftarrow\Phi$
		\FOR{$t\leftarrow1$ \textbf{to} $T$}
			\STATE Receive instance:  $x_t$
			\STATE Predict ${\hat y_t} = F(x) = \sum\limits_{n = 0}^N {{\omega _n}} {f_n}$  as Eq.(1)
			\STATE Reveal ground truth $y_t$  and Update $\omega_n^t+1$  according to Eq.(2)
			\STATE Concept Drift Detection
			\IF{concept drift is True,}
				\STATE Update network via bilevel optimization
			\ELSE
				\STATE Update network via single backpropagation
			\ENDIF
			\STATE Update ${\cal M}^{memry}$  via Reservoir Sample
		\ENDFOR
	\end{algorithmic}
\end{algorithm}
In BODL algorithm, firstly we present an online ensemble framework that at-tempts to dynamically weight the different depth classifiers and the base classifier’s weights for each hidden layer are update based on the exponential gradient descent algorithm in an online manner. In particular, we impose the similar constraint be-tween the shallower and the deeper layer’s features, which would be beneficial to generate the favorable feature representations and improve the performance of the convergence. \\
In addition, consider that the data probability distribution would change in real-world scenarios. Thus, a concept drift detection mechanism is used to monitor the data changes according to the error rate of classifier. Once the drift is detected, the learner would update the model parameters via bilevel optimization. Thus, the learner would effectively prevent the large drift and alleviate the catastrophic forgetting. 

\section{Experiments}
In this section, we evaluate the baselines and our proposed BODL algorithm on vari-ous stationary and non-stationary datasets. We report and analysis the experimental results in detail.
\subsubsection{Experiment Setup}
We use the neural network architecture with 15 hidden layers of 30 units with ReLU nonlinearities. In all experiments, the entire network parameters are updated by Ad-am optimizer with a learning rate of 0.01. When the drift is detected, the model would adaptively learn the parameters via the tiny memory budge and this process is achieved using the bilevel optimization strategy. It is well worth note that we apply a test-then-train strategy for evaluating the learning algorithms to cast this as a classifi-cation task. \\
We compare against with several state-of-the-art baselines: Perceptron, the Re-laxed Online Maximum Margin(ROMMA)\cite{15li2002relaxed}, OGD\cite{16zinkevich2003online}, the recently proposed Soft Confidence Weighted algorithms(SCW) \cite{17hoi2012exact}, the Adaptive Regularization of Weight Vectors(AROW) \cite{18crammer2013adaptive}, the Confidence-Weighted (CW) learning algorithm[19]. Here, the BODL-Base algorithm is regarded as an online learning approach without the bilevel optimization strategy.

\subsubsection{Datasets}
The learning performance of BODL algorithm is numerically validated on stationary and non-stationary data, but evolving data stream usually characterize non-stationary properties in real-word task. Thus, in our experiments, we select three non-stationary datasets and two stationary datasets for experimental comparison. Here, the datasets are obtained from UCI repositories and the properties are shown in de-tails in Table1.
\begin{table}
	\centering
	\caption{Batch Datasets Properties.}
	\label{tb1}
	\begin{tabular}{|l| l| l| l|}\hline
	Dataset & Size & Features & Type \\
	\hline
	MNINST & 70000 & 786 & Stationary \\
	Magic &	19020 & 10	& Stationary \\
	PIMA	& 768 & 	8 &	Non-stationary \\
	Weather &	18140	 & 8 &	Non-stationary \\
	KDDCUP &	1036241 &	127 &	Non-stationary \\
	\hline
	\end{tabular}
\end{table}
\subsubsection{Experimental Results}
In this section, the experimental comparative results of all baselines and the proposed BODL algorithm with four different metric criteria: average accuracy, average preci-sion, F1-Score and recall-score are reported in Table 2. In additional, in order to study the contribution of each component, a complete ablation studies are conducted in our work where BODL-2: the model is trained using the bilevel learning and the simi-lar constrain, BODL-1: the model is trained using the similar constrain alone, BODL-Base: the model is trained without the bilevel learning and the similar constrain.
\begin{table}
	\centering
	\caption{Numerical results of different algorithms on different datasets.}
	\label{tb2}
	\begin{tabular}{|c|c|c|c|c|c|}
		\hline
		\multirow{2}{*}{Method} &\multicolumn{5}{c}{Average Accuracy}\\
		\cline{2-6}& MNIST & Magic &	PIMA &	Weather &	KDDCUP\\
		\hline
		\textbf{BODL-2}	& 92.00\% &	78.73\% &	74.36\% &	74.90\% &	99.68\%\\
		\textbf{BODL-1} &	91.99\% &	78.49\% &	73.84\% &	73.28\% &	99.44\%\\
		\textbf{BODL-Base} &	90.80\% &	78.31\% &	71.69\% &	72.34\% &	99.35\%\\
		Perceptron&	84.77\% &	70.60\% &	64.45\% &	65.85\% &	99.31\%\\
		ROMMA&	83.22\% &	66.67\% &	64.45\% &	65.63\% &	99.34\%\\
		OGD	&90.10\% &	78.72\% &	72.78\% &	72.70\% &	99.61\%\\
		SCW	&88.98\% &	78.64\% &	70.31\% &	76.12\% &	99.75\%\\
		AROW	&89.04\% &	78.71\% &	72.14\% &	75.15\% &	99.58\%\\
		CW	&86.88\% &	67.90\% &	63.41\% &	36.81\% &	99.62\%\\
		PA	&85.68\% &	70.13\% &	66.41\% &	65.74\% &	99.41\%\\
		\hline
		\multirow{2}{*}{Method} &\multicolumn{5}{c}{Average precision}\\
		\cline{2-6}& MNIST & Magic &	PIMA &	Weather &	KDDCUP\\
		BODL-2 &	91.91\% &	74.65\% &	54.83\% &	77.35\% &	98.55\%\\
		BODL-1 &	91.89\% &	73.96\% &	54.38\% &	75.76\% &	97.46\%\\
		BODL-Base &	90.70\% &	 73.80\% &	51.88\% &	76.06\% &	96.99\%\\
		Perceptron &	84.61\% &	67.77\% &	64.03\% &	60.78\% &	98.96\%\\
		ROMMA	 &82.99\% &	64.16\% &	63.74\% &	60.69\% &	99.08\%	\\
		OGD &	89.99\% &	77.66\% &	71.77\% &	67.91\% &	99.35\%\\
		SCW &	88.83\% &	76.75\% &	69.18\% &	74.90\% &	99.55\%\\
		AROW &	88.92\% &	77.62\% &	70.84\% &	71.13\% &	99.31\%\\
		CW &	86.70\% &	64.75\% &	62.55\% &	53.55\% &	99.56\%\\
		PA	 &85.45\% &	67.19\% &	65.06\% &	59.95\% &	99.01\%\\
		\hline
		\multirow{2}{*}{Method} &\multicolumn{5}{c}{F1-score}\\
		\cline{2-6}& MNIST & Magic &	PIMA &	Weather &	KDDCUP\\
		\hline
		BODL-2 &	91.97\% &	65.81\% &	60.83\% &	82.62\% &	99.21\%\\
		BODL-1 &	91.97\% &	65.73\% &	63.30\% &	81.80\% &	98.61\%\\
		BODL-Base &	90.78\% &	65.70\% &	59.66\% &	81.69\% &	98.38\%\\
		Perceptron	 &84.78\% &	70.61\% &	65.27\% &	66.07\% &	99.31\%\\
		ROMMA	 &83.21\% &	67.05\% &	65.27\% &	65.95\% &	99.33\%\\
		OGD &	90.08\% &	78.02\% &	73.39\% &	71.04\% &	99.61\%\\
		SCW &	88.97\% &	78.41\% &	70.96\% &	77.53\% &	99.75\%\\
		AROW &	88.98\% &	78.16\% &	72.73\% &	74.28\% &	99.66\%\\
		CW &	86.87\% &	67.86\% &	64.24\% &	29.00\% &	99.67\%\\
		PA &	85.66\% &	70.08\% &	67.12\% &	65.58\% &	99.41\%\\
		\hline

	\end{tabular}
\end{table}
The experiment results show that our BODL-2 algorithm enjoys competitive per-formance on different datasets implementing different evaluation criteria. BODL-2 is slightly better than BODL-1 with the help of bilevel learning since it can alleviate the catastrophic forget when the concept drift occurs. BODL-Base have lower accuracy than BODL-1, which means the similar constrain would be beneficial to generate the favorable feature representations.\\
Compared to the state-of-the-art methods, we can draw several conclusions. In terms of average accuracy, first but not surprise, traditional online learning tech-niques, such as Perceptron and CW, achieve relatively poor performance on almost all datasets. Next, we also note that the algorithms, such as OGD, could obtain relatively competitive numerical results on MNIST datasets. However, lacked the ability to further explore the power of depth or adaptively adjust the model parameters when concept drift occurs, so they receive poor performance on weather and PIMA da-taset. SCW and AROW achieve favorable accuracy in concept drift datasets such as weather and KDDCUP, but they product poor results in PIMA dataset which features highly imbalance and non-stationary. In contrary, our BODL-2 algorithm can exploit the different-level favorable feature representation base on the deep learning frame-work, besides, when the concept drift is observed, the learner can adaptively adjust the model parameters via bilevel optimization strategy based on memory replay and then encourage positive transfer and prevent the large drift. \\
In additional, BODL-2 algorithm outperform all other approaches on Magic, MNIST and KDDCUP dataset under accuracy evaluation criteria. It is noted that our method can produce good performance from highly imbalance data streams with concept drift, such as PIMA. Only 1.22\% less than the highest one in terms of accu-racy on weather dataset but achieve highest results under the average precision, F1-Score and recall-score evaluation criteria and so on. To conclude, the experimental results demonstrate that our BODL-2 algorithm is a promising online learning ap-proach comparing to the state-of-the-art online methods.

\section{Related Works}
Recent years we have witnessed enormous success in the deep neural network. Com-pared to traditional off-line learning, online learning is more suitable in many real-word tasks. Online learning algorithms represents a class of scalable algorithms which are devised to optimize the models incrementally where the data instance comes gradually. Perceptron based on maximum-margin classification is the earliest online learning algorithm, which is primarily developed to learn linear models. However, the class of perceptron algorithm is fragile to the samples that are linearly inseparable. Thus, perceptron algorithm with the kernel functions are developed \cite{20kivinen2004online}, which give a solution to online learning techniques with nonlinear models. While such approaches are able to solve the non-linear classification, determining the type and number of kernel function is an open challenge. Moreover, these approaches are not explicitly built to extract the different-depth feature representations for the data instances. Base on this fact, Sahoo et al et al present an online algorithm with different depth network for evolving data streams\cite{4sahoo2018online}. However, they neglect the intractable problem of catastrophic forgetting, or cannot cope with the non-stationary environment very well. Recently, there are some specific algorithms handle for concept drift in non-stationary environment. These methods concentrate on incrementally update the model as long as the data instance arrives in a stream, such as dynamic combination model; the online Gradient Descent Algorithm(OGD)\cite{16zinkevich2003online}; the relaxed online maxi-mum margin algorithm and its aggressive version aROMMA, ROMMA, and aROM-MA\cite{15li2002relaxed}; the Adaptive Regularization of Weight Vectors(aROW) \cite{18crammer2013adaptive}; the Confi-dence-Weighted (CW) learning algorithm\cite{19crammer2008exact}; The recently proposed Soft Confidence Weighted algorithms(SCW) \cite{17hoi2012exact}. However, these methods characterize the constant updating of their models, which would make the model evolve in an extremely regu-lar manner regardless of the concept drift.
\section{Conclusion and future work}
Concept drift is an inevitable problem with learning from evolving data streams, which must be handled for data instances to be practically useful. In this work, we proposed a novel Bilevel Online Deep Learning (BODL) framework to learn in non-stationary environment in an online manner. BODL creates an ensemble classifier using the different depth feature representations, where the important weights of each classifier would be updated by online exponential gradient descent strategy. In order to make the deeper layers converge faster and generate the favorable feature repre-sentation, we impose the similar constraint between the shallower and the deeper layer’s features. Besides, a concept drift detected mechanism is devised according to the error rate of classifier. When the concept drift is detected, our BODL algorithm can adaptively update the model parameters via bilevel optimization based on tiny episodic memory and then prevent the large drift and encourage positive transfer. \\
At last, we validated the proposed BODL algorithm through extensive experiments on various stationary and non-stationary datasets and the competitive numerical results show our BODL algorithm is a promising online learning approach.\\
In the future work, we would consider the online learning problem for class incre-mental learning. Besides, in order to obtain the more favorable feature representa-tion, we also consider incorporating the recently proposed self-supervised learning and data augment methods.

\section{Acknowledgements}
This work was supported by the Science Foundation of China University of Petroleum, Beijing(No. 2462020YXZZ023)
\bibliography{Bilevel}
\end{document}